\documentclass[11pt]{article}

\usepackage[final]{acl}

\usepackage{times}
\usepackage{latexsym}
\usepackage{amssymb}
\usepackage{amsmath}
\usepackage{booktabs}
\usepackage{subcaption} 
\usepackage[most]{tcolorbox}
\colorlet{promptbg}{blue!15}
\usepackage[T1]{fontenc}

\usepackage[utf8]{inputenc}

\usepackage{microtype}

\usepackage{inconsolata}

\usepackage{graphicx}

\title{DWA-KD: Dual-Space Weighting and Time-Warped Alignment for Cross-Tokenizer Knowledge Distillation}

\renewcommand{\thefootnote}{\fnsymbol{footnote}}

\author{
 \textbf{Duc Trung Vu\textsuperscript{1}\footnotemark[1]},
 \textbf{Pham Khanh Chi\textsuperscript{1}\footnotemark[1]},
 \textbf{Dat Phi Van\textsuperscript{1}\footnotemark[1]},
 \textbf{Linh Ngo Van\textsuperscript{1}},
\\
 \textbf{Sang Dinh\textsuperscript{1}\footnotemark[2]},
  \textbf{Trung Le\textsuperscript{2}}
\\
 \textsuperscript{1} Hanoi University of Science and Technology, Vietnam,\\
  \textsuperscript{2} University of Monash, Australia,}

\begin{document}
\maketitle

\begingroup
\renewcommand{\thefootnote}{*}
\footnotetext{Equal contribution.}
\renewcommand{\thefootnote}{$\dagger$}
\footnotetext{Corresponding author: sangdv@soict.hust.edu.vn}
\endgroup

\begin{abstract}
Knowledge Distillation (KD) has emerged as a crucial technique for compressing Large Language Models (LLMs). Although existing cross-tokenizer KD methods have made notable progress, their effectiveness remains constrained by suboptimal alignment across sequence and vocabulary levels. To address these limitations, we introduce Dual-Space Weighting and Time-Warped Alignment (DWA-KD), a novel cross-tokenizer distillation framework that enhances token-wise distillation through dual-space entropy-based weighting and achieves precise sequence-level alignment by leveraging both lexical and semantic information. At the token level, DWA-KD maps teacher representations into the student space and vice versa, performing dual-space KD via Kullback–Leibler divergence (KL). The process is modulated by dual-space weights that up-weight tokens where the student is uncertain and the teacher is confident, thereby focusing learning on informative tokens rather than treating all positions equally. At the sequence level, DWA-KD applies Soft Dynamic Time Warping (Soft-DTW) to both the embedding and final hidden-state layers, enabling robust alignment of lexical and contextual semantics between teacher and student sequences. Extensive experiments across diverse NLP benchmarks demonstrate that DWA-KD outperforms state-of-the-art KD baselines, while ablation studies confirm the complementary contributions of entropy-based token weighting and embedding and final hidden state layer Soft-DTW alignment.
\end{abstract}

\section{Introduction}

Large language models deliver strong results but incur high compute, memory, and latency~\cite{kaplan2020scaling, chowdhery2023palm, touvron2023llama2, openai2023gpt4}. This creates a need for smaller models that remain efficient in practice. Knowledge distillation (KD) \cite{hinton2015distilling} has emerged as a promising solution to this challenge by transferring knowledge from a large teacher model to a compact student model, maintaining essential performance while reducing inference overhead \citep{ko2024distillm,chen2025cdm,gu2024minillm}.
Conventional KD is effective for LLMs but is often constrained by a \emph{same-tokenizer} setting, i.e., requiring shared vocabulary between teacher and student, which limits applicability across heterogeneous models \cite{wen2023fdivergence, gu2024minillm, ko2024distillm}.

Cross-tokenizer KD (CTKD) removes the shared-tokenizer assumption but introduces main challenges: (i) \emph{tokenization mismatch} and (ii) \emph{sequence misalignment}. Recent CTKD methods address these challenges by leveraging Optimal Transport (OT)~\citep{boizard2024cross, cui2025multilevelot}, learning projectors and uses cross-model attention to reconcile hidden spaces~\citep{zhang2024dual} or aligning strings with dynamic programming on token sequences~\citep{wan2024knowledge}. While OT-based sequence alignment \citep{boizard2024cross, cui2025multilevelot, truong-etal-2025-emo} does not inherently preserve temporal order~\citep{su2017order}, \citet{wan2024knowledge} and \citet{zhang2024dual} only captures surface form or logit-level and overlook semantics \citep{chen2025cdm}. Meanwhile, a novel method - Contextual Dynamic Mapping (CDM)~\citep{chen2025cdm} - introduces entropy-weighted dynamic programming based on Dynamic Time Warping (DTW) to adapt token alignment.
However, CDM is computationally expensive. Its string-level DP uses CPU-side loops each forward pass, which leads to GPU under-utilization. Recent work has also explored cross-tokenizer distillation under preference/alignment signals \citep{nguyen2026ctpd}.

DTW-based objectives also appear in broader settings (e.g., ~\citet{fu2023specializing, wan2024knowledge}), which is used to align at sequence-level.  Despite recent progress in cross-tokenizer knowledge distillation, current approaches exhibit significant limitations. These methods assume all tokens contribute equally, neglecting the varying importance or informativeness of individual tokens. Moreover, sequence-level context is often under-weighted in CTKD, weakening robustness when tokenizers diverge substantially.

To address these limitations, we propose \textbf{Dual-Space Weighting and Time-Warped Alignment (DWA-KD)}, a CTKD framework that aligns teacher and student at both token and sequence levels. 
At the token level, we adopt an asymmetric scheme: on the \emph{student side}, each position is weighted by the product of the student's normalized entropy and the teacher's \emph{projected} confidence (the maximum probability after mapping the teacher distribution into the student vocabulary), emphasizing updates where the student is uncertain and the teacher target is reliable; on the \emph{teacher side}, we use \emph{entropy-only} weighting, assigning larger weights to low-entropy (confident) teacher tokens. 
This dual mechanism concentrates learning on the most consequential tokens while avoiding noisy or redundant updates. 
At the sequence level, we employ normalized Soft-DTW~\citep{cuturi2017softdtw} with an attention-informed soft band over both embeddings and final hidden states, reducing input-level mismatch and encouraging the student to track the teacher’s semantic trajectory before producing logits.
 
We evaluate our framework across multiple benchmarks under cross-tokenizer setting. The experimental results demonstrate that DWA-KD exhibits superiority over existing cross-tokenizer distillation approaches among different datasets. Furthermore, ablation study confirms the effectiveness of our main components. In summary, the main contributions of this work are:
\begin{itemize}
  \item We present \textbf{Dual-Space Weighting and Time-Warped Alignment (DWA-KD)}, a cross-tokenizer distillation framework that couples selective \textit{token-level} transfer with \textit{sequence-level} Soft-DTW, applied to both embeddings and final hidden states, with an \textit{attention-informed soft band} that keeps alignments plausible and gradients smooth.
  \item We introduce \textbf{a dual-space token weighting scheme}: on the student side, weights scale with the product of student uncertainty and teacher $\to$ student projected confidence; on the teacher side, weights increase with teacher confidence (low entropy). This focuses learning on tokens that matter while suppressing noisy updates.
  \item We conduct comprehensive experiments across diverse datasets to evaluate our approach, showing that it enhances the performance of student models compared to standard knowledge distillation baselines. 
\end{itemize}

\section{Related Work and Background}
\paragraph{Knowledge Distillation}

Knowledge distillation (KD) is a widely adopted technique for transferring knowledge from a teacher to a student model by aligning soft targets such as logits or intermediate representations \cite{hinton2015distilling, kim2016sequence}. With the rise of large language models (LLMs), many conventional KD methods \cite{park2021distilling, gu2023minillm, wu2024rethinking, ko2024distillm}, while effective, are often limited to same-tokenizer setting. Cross-tokenizer KD \cite{zhou2022bert, zhang2023reaugkd, liu2022multigranularity} addresses this challenge but introduce other problem, mainly tokenization mismatch and sequence mismatch. 
To address these issues, DSKD~\citep{zhang2024dual} aligns tokens by learning projectors that map student and teacher hidden states into a shared space, but it treats all positions equally and ignores token-wise importance. MultiLevelOT~\citep{cui2025multilevelot} and MCW-KD~\citep{vuong2026mcwkd} leverage Wasserstein/OT-style objectives for distillation.
Beyond token-level matching, MultiLevelOT~\citep{cui2025multilevelot} and CDM~\citep{chen2025cdm} further extend alignment to token and sequence levels: MultiLevelOT measures discrepancies within and across tokens via optimal transport but does not inherently preserve temporal order, while CDM applies entropy-weighted DTW in the logit space, which can underuse contextual information encoded in hidden states.
 In contrast, our approach employs asymmetric token weighting and applies normalized Soft-DTW at both the embedding and final hidden-state layers. This combination reduces input-level mismatch and better captures semantic trajectories, enabling more precise and efficient knowledge transfer.
\paragraph{Dynamic Time Warping}

Dynamic Time Warping (DTW) \cite{sakoe1978dtw} is a classical algorithm for computing optimal alignments between temporal sequences, accommodating differences in length or variations in speed. Its original formulation, however, is non-differentiable, limiting use in gradient-based learning. Soft Dynamic Time Warping (Soft-DTW) \cite{cuturi2017softdtw} addresses this by replacing the hard minimum with a smooth approximation, enabling end-to-end training. Soft-DTW has demonstrated improved stability and performance in domains such as time series regression, speech, handwriting, and motion trajectory prediction. Building on this, several DTW-based methods leverage soft alignment to capture temporal or structural correspondences between teacher and student representations, aligning token spans before pooling logits for divergence computation \cite{fu2023specializing, wan2024knowledge, chen2025cdm}. In contrast, our approach applies Soft-DTW directly at the embedding and hidden-state levels, providing differentiable alignment that captures both lexical and semantic dependencies for more effective knowledge transfer.
\subsection{Background}
\subsubsection{Dual-Space Knowledge Distillation}

DSKD \citep{zhang2024dual} addresses tokenization mismatch by learning soft correspondences between student and teacher tokens via a cross-model attention (CMA) layer. Concretely, student embeddings of input and target tokens are concatenated and projected to query vectors \(Q\in\mathbb{R}^{S\times 2D}\); teacher embeddings and output hidden states are then normalized to form key vectors \(K\in\mathbb{R}^{T\times 2D}\) and projected to value vectors \(V\in\mathbb{R}^{T\times d}\). Scaled dot-product attention is formalized as:
\begin{equation}
a^{t \to s} = \mathrm{softmax}\!\left(\frac{QK^{\top}}{\sqrt{2D}}\right) \in \mathbb{R}^{S \times T}.
\end{equation}
produces teacher-to-student alignments and being used to align teacher hidden state to student hidden state \(\tilde{h}^{t \to s} = a^{t \to s} V \in \mathbb{R}^{n \times d}.\) A reverse alignment matrix is defined as:
\begin{equation}
a^{s \to t} = \text{softmax} \left( \frac{KQ^\top}{\sqrt{2D}} \right) \in \mathbb{R}^{T \times S}
\end{equation}
and also be used to project and align student’s hidden states to teacher’s hidden state \(
\tilde{h}^{s \to t} = a^{s \to t} P^{s \to t}
\big(\, h^s ; \theta^{s \to t}_P \,\big) \in \mathbb{R}^{T \times D}.
\) where $h^s$ is the output hidden states of the whole sequence from the student model.

\paragraph{Dual-Space KD.}
(i) \emph{Student space:} \(\tilde{h}^{t \to s}\) is mapped by the student head to produce output distribution \(p^{t\to s}\).
Because the projectors start at random, an auxiliary cross-entropy is used to warm them up; then a KL term matches \(p^{t\to s}\) to the student distribution \(p^s\).
(ii) \emph{Teacher space:} reverse-aligned hidden state $\tilde{h}^{s \to t}$ are passed through the teacher head to obtain \(p^{s\to t}\), which is aligned to the teacher distribution \(p^t\) via KL.
\paragraph{Objective.}
The final loss sums KD in both spaces plus the auxiliary CE:
\begin{equation}
\mathcal{L}_{\mathrm{DSKD}}
=\mathcal{L}_{\mathrm{kd}}^{\mathrm{stu}}
+\mathcal{L}_{\mathrm{kd}}^{\mathrm{tea}}
+\mathcal{L}_{\mathrm{ce}}^{t\to s}.
\label{eq:dskd_loss}
\end{equation}

\subsubsection{Dynamic Time Warping (DTW)}

DTW~\citep{sakoe1978dtw} is a dynamic-programming method that aligns two time sequences by allowing non-linear time warping. This property makes DTW particularly useful for sequences with varying lengths or speed variations. Formally, given sequences
\(
X = (x_1, \dots, x_N), \) and \(
Y = (y_1, \dots, y_M),
\)
DTW constructs a cost matrix $\mathbf{C} \in \mathbb{R}^{N \times M}$ with entries:
\begin{equation}
C_{ij} = d(x_i, y_j),
\end{equation}
where $d(\cdot, \cdot)$ is a local distance function (e.g., squared Euclidean distance), and computes the accumulated cost matrix $\mathbf{R}$ recursively via:
\begin{equation}
R_{ij} = C_{ij} + \min \big( R_{i-1,j}, \, R_{i,j-1}, \, R_{i-1,j-1} \big).
\end{equation}
While DTW preserves the temporal order of elements, the non-differentiability of the $\min$ operator prevents its direct integration into gradient-based learning frameworks. Soft Dynamic Time Warping (Soft-DTW) \cite{cuturi2017softdtw} extends DTW by introducing a differentiable relaxation of the alignment cost. The non-differentiable $\min$ in the recurrence is replaced with a smooth softmin operator:
\begin{equation}
    \min{}^\gamma(a_1, \dots, a_n) = - \gamma \log \sum_i e^{-a_i / \gamma}
\end{equation}
where $\gamma > 0$ controls the smoothness and as $\gamma \to 0$, it converges to the standard minimum. Incorporating this operator into the DTW recurrence yields a differentiable cost that supports backpropagation, enabling end-to-end training in sequence modeling tasks such as knowledge distillation, speech modeling, and time series prediction.

\section{Methodology}

\subsection{Notations}
Let the student and teacher models produce token sequences 
$\mathbf{x} = (x_1, x_2, \ldots, x_S)$ and 
$\mathbf{y} = (y_1, y_2, \ldots, y_T)$, 
with respective vocabularies of sizes $V^s$ and $V^t$. 
The student’s predictive distribution over its vocabulary is denoted as 
\(p^s(x_i \mid x_{<i}) \in \mathbb{R}^{V^s}\), and the teacher’s predictive distribution over its own vocabulary is denoted as 
\(p^t(y_j \mid y_{<j}) \in \mathbb{R}^{V^t}\). The entropy of the student model at position $i$ is defined as
\begin{equation}
H^s(x_i) = - \sum_{i=1}^{V^s} p^s(x_i \mid x_{<i}) \log p^s(x_i \mid x_{<i}),
\end{equation}
and the entropy of the teacher model at position $j$ is given by
\begin{equation}
H^t(y_j) = - \sum_{j=1}^{V^t} p^t(y_j \mid y_{<j}) \log p^t(y_j \mid y_{<j}).
\end{equation}
Additionally, let the student and teacher produce embedding sequences 
\( E^{s} = (e^s_1, e^s_2, \ldots, e^s_S) \in \mathbb{R}^{S \times d_S} \) and \(E^{t} = (e^t_1, e^t_2, \ldots, e^t_T) \in \mathbb{R}^{T \times d_T},
\)
as well as final hidden-state representations
\( H^{s} = (h^s_1, h^s_2, \ldots, h^s_S) \in \mathbb{R}^{S \times d_S} \) and \( H^{t} = (h^t_1, h^t_2, \ldots, h^t_T) \in \mathbb{R}^{T \times d_T}.\)
Since teacher and student models typically differ in dimensionality, two lightweight linear projections 
$W^{t \rightarrow s}_e, W^{t \rightarrow s}_h \in \mathbb{R}^{d_S \times d_T}$ 
are introduced to map teacher representations into the student space:
\begin{align}
\tilde{E}^{t} &= E^{t}(W^{t\rightarrow s}_e)^\top, \label{eq:proj_e}\\
\tilde{H}^{t} &= H^{t}(W^{t\rightarrow s}_h)^\top. \label{eq:proj_h}
\end{align}
Moreover, as in DSKD, the teacher to student distribution $p^{t \to s}(x_i \mid x_{<i})$ is aligned to the student distribution $p^s(x_i \mid x_{<i})$ via $D(\cdot \mid \cdot)$. Conversely, student to teacher distribution $p^{s \to t}(y_j \mid y_{<j})$ is aligned to the teacher distribution $p^t(y_j \mid y_{<j})$ via 
$D(\cdot \mid \cdot)$. Here $D(\cdot \mid \cdot)$ denotes KL divergence.

\subsection{Weighted Dual-Space Knowledge Distillation}
In sequence modeling, not all tokens contribute equally to the learning process. Certain tokens carry more informative or uncertain content, whereas others are predictable or less critical \cite{zhong2024revisiting, akhauri2025tokenbutler}. To more effectively guide the student model during knowledge distillation, it is beneficial to assign distinct weights to individual tokens, thereby emphasizing those of greater significance \citep{le2025tokenlevelselfplay}. Our approach enables the learning process to concentrate on the most certain teacher token  elements of the sequence, enhancing both training efficiency and overall model performance. In this work, we propose a mechanism for assigning weights to each token in both the student and teacher spaces. To ensure stability during training, we scale the token weights so that their sum equals the sequence length. This normalization makes the KD process robust to both sequence length variability and shifting entropy distributions as the model learns, avoiding the need for constant retuning of loss coefficients.

\paragraph{Token weight in student space.}

Inspired by \citet{wang2025highentropy} and  \citet{furlanello2018born}, in the student space, we weight each token by combining the student's uncertainty with the teacher's confidence projected into the student vocabulary. Formally, we define peak confidence from teacher at student sequence position $i$ as:

\begin{equation}
g^{t \to s}(x_i) \;=\; \max_{i \in V_s} \; p^{t \to s}(x_i \mid x_{<i}).
\label{eq:gate_teacher_to_student}
\end{equation}
The weight $w_i^{(s)}$ of student token $i$ is defined as:
\begin{equation}
w^{(s)}_i = {H}^{s}(x_i)\;\cdot\; g^{t\to s}(x_i).
\label{eq:stu-weight-mul}
\end{equation}
By weighting each student token as Eq.~\ref{eq:stu-weight-mul}, we consider a token to be important only when the student needs help (high uncertainty), and that token has a reliable teacher target in the student’s vocabulary (high peak confidence).
Consequently, the knowledge distillation loss in the student space is formulated as:
\begin{equation}
\widetilde{L}_{\text{KD}}^{\text{stu}} = \sum_i w_i^{(s)} \, D(p^{t \to s}(x_i \mid x_{<i}) \;\|\; p^s(x_i \mid x_{<i} )),
\end{equation}
This weighting mechanism prevent wasting capacity on tokens the student already knows (low entropy) or where the teacher is unsure (low max-probability). 

\paragraph{Token weight in teacher space.}

In the teacher space, we assign token weights based on the teacher’s confidence, computed as a normalized complement of the token-level entropy. 



\begin{equation}
w^{(t)}_j = 1 - \frac{H^{t}(y_j)}{\log V^t}
\end{equation}

This weighting mechanism ensures that the student primarily learns from tokens where the teacher is confident (low entropy), while ignoring uncertain teacher tokens, thereby reducing potential noise. Consequently, the knowledge distillation loss in the teacher space is formulated as:

\begin{equation}
\widetilde{L}_{\text{KD}}^{\text{tea}} = \sum_j w_j^{(t)} \, D\big(p^t(x_j \mid x_{<j}) \,\|\, p^{s \to t}(x_j \mid x_{<j})\big)
\end{equation}

\paragraph{Token alignment loss.}The overall loss of token level alignment combines the KD losses in both spaces with the cross-entropy loss:
\begin{equation}
    {L}_{\text{W-KD}} = \widetilde{L}_{\text{KD}}^{\text{stu}} + \widetilde{L}_{\text{KD}}^{\text{tea}} + {L}_{\text{CE}}^{t \to s}.
\end{equation}

\subsection{Dynamic Time Warping Alignment}
We align teacher and student embeddings and hidden states in shared latent spaces via DTW loss to transfer representational and temporal knowledge across models with different architectures.

\paragraph{Pairwise costs.}
For each representation type (embedding or hidden), we measure pairwise dissimilarity between the student sequence and the projected-teacher sequence using cosine distance.\footnote{We share the projector for hidden-state representations with the projector used in DSKD and introduce an independent projector for embeddings.} Let $S$ and $T$ be the student/teacher sequence lengths. The cost matrices $C \in \mathbb{R}^{S \times T}$ are
\begin{align}
C\!\big(E^{\text{stu}}, \tilde{E}^{\text{tea}}\big)
&= 1 - \cos\!\big(E^{\text{stu}}, \tilde{E}^{\text{tea}}\big), \label{eq:cost_matrix_E}\\
C\!\big(H^{\text{stu}}, \tilde{H}^{\text{tea}}\big)
&= 1 - \cos\!\big(H^{\text{stu}}, \tilde{H}^{\text{tea}}\big), \label{eq:cost_matrix_H}
\end{align}
where $\cos(\cdot,\cdot)$ denotes cosine similarity applied row-wise to produce an $S\times T$ matrix.
\paragraph{Banding via Attention Entropy.} We start from the Sakoe–Chiba (SC) idea \citep{sakoe1978dtw} and discourage off‑diagonal warps by adding a fixed penalty to cross‑cost entries outside a diagonal band. Rather than a hard mask, this soft penalty keeps the objective differentiable. The band is adapted per student position using DSKD cross‑model attention matrix A between student and teacher tokens. Let $A \in \mathbb{R}^{\text{S}\times \text{T}}$ be the row‑normalized cross‑model attention from student tokens to teacher tokens (rows sum to 1). We apply a row‑wise softmax over teacher tokens, and use that distribution to set the band. For each student index, we form a soft center as the attention‑weighted average teacher index, a linear center from the proportional diagonal, and blend the two to obtain the row’s band center; the band width widens with the row’s attention entropy. Formally, for each row $i$: 

\begin{equation}
c_i = \alpha \sum_{j=1}^{T} j A_{i,j} + (1-\alpha) i \cdot \frac{T}{S}.
\label{eq:center}
\end{equation}
The band width is then set from the attention entropy - high entropy (uncertain attention) yields a wider band, while low entropy tightens it:
\begin{equation}
H_i = - \sum_{j=1}^{T} A_{i,j} \log A_{i,j},
\end{equation}
And the adaptive width is the combination of base width $b$ (the minimum band size even under low uncertainty) and attention entropy $H_i$:

\begin{equation}
w_i = b + \beta H_i.
\end{equation}
The final cross-cost is penalized outside the adaptive band:

\begin{equation}
\tilde{C}^{s,t}_{i,j} = C^{s,t}_{i,j} + \lambda_{\text{band}} \cdot \mathbf{1}_{|j - c_i| > w_i}.
\end{equation}
Additional details of formulation of attention matrix $A$ are provided in Appendix \ref{sec:appendix_D}.
\paragraph{Sequence-level Alignment via Soft-DTW}
Let $s_{\gamma}(X,Y) \equiv \mathrm{sDTW}_{\gamma}\!\big(\widetilde{C}(X,Y)\big)$ denote the Soft-DTW score on the cost matrix with adaptive band $\widetilde{C}(X,Y)$. Following prior work on \emph{Soft-DTW divergence} \citep{cuturi2017softdtw,blondel2021differentiable}, we remove self-similarity bias and enforce symmetry by
\begin{align}
\mathrm{nDTW}_{\gamma}(X,Y) = s_{\gamma}(X,Y) - \Delta_{\gamma}(X,Y)\\
\Delta_{\gamma}(X,Y) \;=\; \tfrac{1}{2}\!\left[s_{\gamma}(X,X)+s_{\gamma}(Y,Y)\right].
\label{eq:ndtw}
\end{align}
We apply these formulas to the pairs $(X,Y)\in\{(E^{\text{stu}},\tilde{E}^{\,\text{tea}}),\; (H^{\text{stu}},\tilde{H}^{\,\text{tea}})\}$. This normalization makes distances comparable across sequences of different lengths or embedding magnitudes and stabilizes gradients. We sum the divergences at the hidden and embedding levels:
\begin{align} 
L_{\text{DTW}} \;=\; nDTW_\gamma\!\big(H^{stu},\,\tilde{H}^{\,tea}\big) \nonumber \\ + nDTW_\gamma\!\big(E^{stu},\,\tilde{E}^{\,tea}\big) 
\label{eq:dtw_loss} 
\end{align}
The first term promotes temporal agreement between hidden-state trajectories; the second anchors lexical correspondences via token embeddings. Together, they encourage the student to absorb both contextual (semantic) and lexical structure under tokenizer and length mismatches.

\subsection{Overall Loss}

The final training objective is formulated as a weighted combination of complementary components:
\begin{equation*}
L = \lambda_{\text{CE}} L_{\text{CE}}
            + \lambda_{\text{W-KD}} L_{\text{W-KD}}
            + \lambda_{\text{DTW}} L_{\text{DTW}} 
\label{eq:final-loss}
\end{equation*}
Here, $\mathcal{L}_{\text{CE}}$ denotes the standard cross-entropy loss for supervised learning, 
$\mathcal{L}_{\text{W-KD}}$ represents the knowledge distillation objective that transfers soft distributions 
from the teacher, and $\mathcal{L}_{\text{DTW}}$ captures temporal alignment between student and teacher via 
dynamic time warping. The coefficients $\lambda_{\text{CE}}, \lambda_{\text{W-KD}}, \lambda_{\text{DTW}}$ balance 
the contribution of each term.

\section{Experiments}
Our experiments demonstrate that DWA-KD outperforms existing cross-tokenizer distillation methods across multiple benchmarks. Ablation studies further verify the effectiveness of entropy-based token weighting and DTW with attention-informed banding in enhancing alignment and knowledge transfer. Additional experiments are in appendix \ref{appdx: additional}.

\subsection{Settings}
\subsubsection*{Baselines and LLMs.}
We evaluate our method in cross-tokenizer settings using both small- and large-scale teacher-student pairs. For the small-scale setup, we consider Qwen 1.5–1.8B as the teacher and GPT-2 120M as the student, and Qwen 1.5–1.8B as the teacher and GPT-2 340M as the student. For the large-scale setup, we use Mistral 7B as the teacher with TinyLlama 1.1B as the student, and Qwen 2.5–7B as the teacher with GPT-2 1.5B and OPT 2.7B as the student. We benchmark our approach against several cross-tokenizer knowledge distillation baselines, including ULD \cite{boizard2024cross}, MinED \cite{wan2024knowledge}, MultiLevelOT \cite{cui2025multilevelot}, and DSKD \cite{zhang2024dual} \footnote[1]{
We exclude CDM~\citep{chen2025cdm} from baselines because its string-level dynamic programming 
requires CPU-side loops at each forward pass, making large-scale GPU training impractical.
}.

\subsubsection*{Datasets}
We perform experiments on multiple instruction-following datasets. Following the approach of \cite{gu2024minillm}, we employ the DATABRICKS-DOLLY-15K dataset for the knowledge distillation process. For evaluation purposes, beyond this primary dataset, we incorporate three instruction-tuning benchmarks as supplementary test sets for out-of-distribution assessment: Super-Natural-Instructions (\textbf{S-NI}) \cite{wang2022benchmarking}, Vicuna Evaluation (VicunaEval) \cite{chiang2023vicuna}, and Self-Instruct (Self-Inst) \cite{wang2023selfinstruct}. This comprehensive evaluation enables us to examine the model’s generalization performance across a broad spectrum of instruction domains.

\subsubsection*{Training and Evaluation}
For GPT-2, we fully fine-tune both teacher and student models, while for TinyLLaMA we apply LoRA fine-tuning. Specifically, the temperature $\tau$ is set to 2.0 based on validation set performance. In addition, all projectors in our approach are implemented as linear layers, which introduce only a small number of additional parameters during training. For evaluation, we generate responses from each model using five different random seeds, and the final performance is quantified by Rouge-L~\cite{lin2004rouge} between the generated outputs and the human-annotated references.
Additional details on model configurations, along with the training and evaluation setup, are provided in Appendix \ref{sec:appendix_B}.

\subsection{Main Results}
\begin{table}[ht]
\centering
\renewcommand{\arraystretch}{1}
\setlength{\tabcolsep}{1pt} 

\begin{tabular}{l|ccccc}
\toprule
\textbf{Methods} & \textbf{Dolly} & \textbf{SelfInst} & \textbf{Vicuna} & \textbf{S-NI} & \textbf{Avg.} \\
\midrule
\multicolumn{6}{c}{\textit{Qwen-1.5$-$1.8B} $\rightarrow$ \textit{GPT2-120M}}\\
\midrule
Teacher        & 28.23 & 19.58 & 19.59 & 34.36 & 25.44 \\
\midrule
SFT            & 23.78 &  6.78 & \textbf{17.04} &  7.81 & 13.85 \\
ULD            & 23.77 &  9.30 & 14.33 & 14.04 & 15.36 \\
DSKD           & 24.26 & 10.07 & 15.25 & 17.15 & 16.68 \\
MinED          & 24.21 & 10.02 & 14.96 & 16.40 & 16.40 \\
MultiLevelOT   & 23.02 &  8.41 & 13.79 & 12.26 & 14.37 \\
\textbf{DWA-KD}& \textbf{24.29} & \textbf{11.15} & 15.64 & \textbf{19.63} & \textbf{17.68} \\
\midrule
\multicolumn{6}{c}{\textit{Qwen-1.5$-$1.8B} $\rightarrow$ \textit{GPT2-340M}}\\
\midrule
Teacher        & 28.23 & 19.58 & 19.59 & 34.36 & 25.44 \\
\midrule
SFT            & 23.11 &  9.09 & 14.89 & 13.03 & 15.03 \\
ULD            & 23.90 &  9.96 & 15.04 & 16.26 & 16.29 \\
DSKD           & 25.43 & 11.29 & 15.08 & 17.18 & 17.25 \\
MinED          & 24.48 & 11.21 & 15.56 & 15.69 & 16.74 \\
MultiLevelOT   & 23.95 & 10.21 & 14.80 & 15.87 & 16.21 \\
\textbf{DWA-KD}& \textbf{26.64} & \textbf{12.10} & \textbf{16.99} & \textbf{21.67} & \textbf{19.35} \\
\bottomrule
\end{tabular}%
\caption{Rouge-L scores (in \%) averaged over five random seeds across multiple benchmarks for two teacher–student pairs.}
\label{tab:icare-main}
\end{table}

\begin{table}[t]
\centering
\renewcommand{\arraystretch}{1}
\setlength{\tabcolsep}{1pt}

\begin{tabular}{l|ccccc}
\toprule
\textbf{Methods} & \textbf{Dolly} & \textbf{SelfInst} & \textbf{Vicuna} & \textbf{S-NI} & \textbf{Avg} \\
\midrule
\multicolumn{6}{c}{\textit{Qwen2.5{-}7B $\rightarrow$ GPT2{-}1.5B}}\\
\midrule
Teacher      & 28.49 & 24.67 & 20.48 & 39.87 & 28.38 \\
\midrule
SFT          & 21.83 & 13.62 & 15.95 & 21.66 & 18.27 \\
ULD          & 24.52 & 15.11 & 15.94 & 26.18 & 20.44 \\
MinED        & 25.52 & 15.39 & 16.15 & 26.25 & 20.83 \\
MultiLevelOT & 24.40 & 14.53 & 15.97 & 23.94 & 19.71 \\
DSKD         & 25.38 & 16.10 & 16.84 & 25.82 & 21.04 \\
\textbf{DWA-KD} & \textbf{26.64} & \textbf{16.54} & \textbf{17.86} & \textbf{30.34} & \textbf{22.85} \\
\midrule
\multicolumn{6}{c}{\textit{Qwen2.5{-}7B{-}Instruct $\rightarrow$ OPT{-}2.7B}}\\
\midrule
Teacher      & 28.49 & 24.67 & 20.48 & 39.87 & 28.38 \\
\midrule
SFT          & 27.10 & 13.90 & 16.60 & 24.90 & 20.63 \\
ULD          & 26.65 & 15.37 & 16.97 & 25.44 & 21.11 \\
MinED        & 26.89 & 14.98 & 17.04 & 25.94 & 21.21 \\
MultiLevelOT & 26.76 & 15.51 & 16.56 & 24.84 & 20.92 \\
DSKD         & 26.93 & \textbf{16.22} & 17.86 & 27.33 & 22.09 \\
\textbf{DWA-KD} & \textbf{28.45} & 15.65 & \textbf{18.16} & \textbf{30.54} & \textbf{23.20} \\
\midrule
\multicolumn{6}{c}{\textit{Mistral-7B} $\rightarrow$ \textit{TinyLLaMA-1.1B}}\\
\midrule
Teacher        & 31.56 & 25.10 & 20.50 & 36.07 & 28.31 \\
\midrule
SFT            & 23.20 & 14.88 & 16.42 & 27.79 & 20.57 \\
ULD            & 25.48 & 17.72 & 17.31 & 32.54 & 23.26 \\
DSKD           & 26.28 & 17.19 & \textbf{18.74} & 31.93 & 23.54 \\
MinED          & 25.54 & 18.23 & 17.02 & 31.42 & 23.05 \\
MultiLevelOT   & 24.56 & 15.61 & 16.84 & 27.91 & 21.23 \\
\textbf{DWA-KD} & \textbf{26.38} & \textbf{19.62} & 17.63 & \textbf{36.59} & \textbf{25.06} \\
\bottomrule
\end{tabular}%
\caption{Rouge-L (\%) on multiple teacher–student pairs. Bold indicates the best \emph{student} score per block.}
\label{tab:qwen25_pairs}
\end{table}

Table~\ref{tab:icare-main} and table~\ref{tab:qwen25_pairs} summarizes the ROUGE-L performance of DWA-KD compared with standard supervised fine-tuning (SFT) and recent state-of-the-art cross-tokenizer knowledge distillation (CTKD) baselines, including ULD, DSKD, MinED, and MultiLevelOT. Evaluations were conducted across four teacher–student model pairs with distinct tokenization schemes on multiple benchmarks (Dolly, Self-Inst, Vicuna, and S-NI).
Overall, DWA-KD achieves the best or competitive ROUGE-L scores across all teacher–student model pairs. For the Mistral-7B → TinyLLaMA-1.1B setting, DWA-KD attains the highest scores on four benchmarks, while remaining close to the best method on Dolly. In the Qwen-1.5B → GPT2-340M pair, it again delivers the best overall results, showing clear improvements on Vicuna and S-NI. For the Qwen2.5-7B-Instruct → GPT2-1.5B configuration, DWA-KD outperforms all baselines, being only slightly below DSKD on SelfInst. These results demonstrate that DWA-KD maintains strong and stable performance across diverse model scales and datasets.
These results confirm that DWA-KD provides consistent improvements over existing CTKD approaches, validating the effectiveness of its dual-space weighting and time-warped alignment mechanisms.

\subsection{Evaluation via GPT-4}
\begin{figure}[t]
    \centering
    \includegraphics[width=1\linewidth]{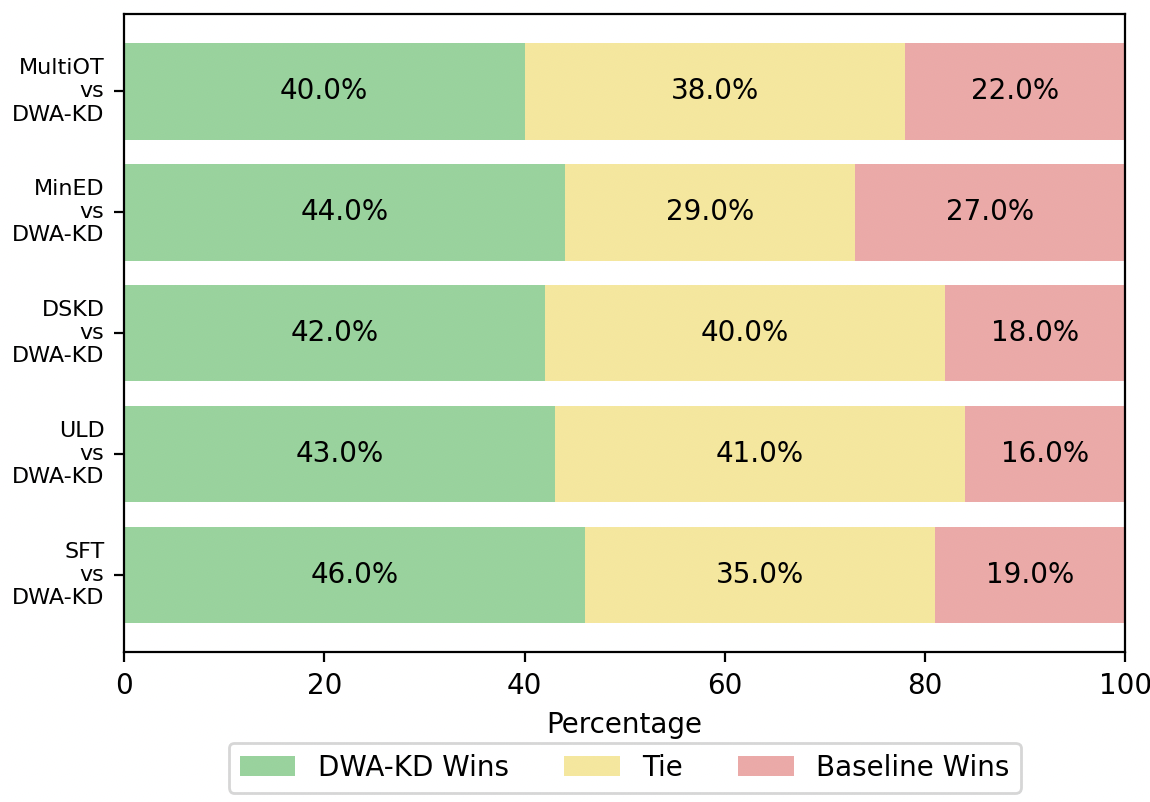}
    \caption{Win rates (\%) for distilling Mistral-7B to TinyLLaMA-1.1B , evaluated by GPT-4.1 on response quality.
    }
    \label{fig:placeholder}
\end{figure}

\begin{figure}[t]
    \centering
    \includegraphics[width=1\linewidth]{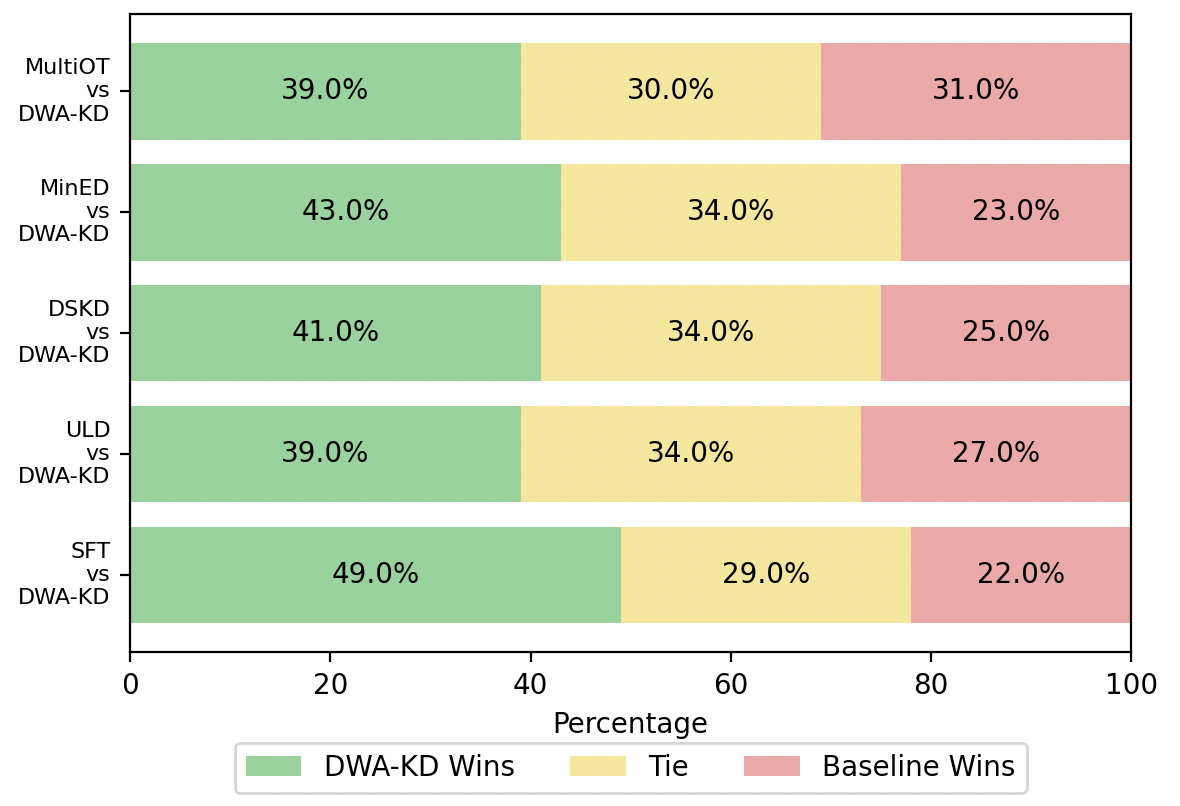}
    \caption{Win rates (\%) for distilling Qwen2.5-7B-Instruct to OPT-2.7B , evaluated by GPT-4.1 on response quality.
    }
    \label{fig:placeholder2}
\end{figure}

To complement ROUGE-L, we also use the \texttt{GPT-4.1} API as an automatic judge for pairwise comparisons between our method and each baseline. With a fixed evaluation prompt (Figure~\ref{fig:gpt4}), the model selects the response that better follows the instruction; to reduce position bias~\citep{zheng2023judging}, outputs are randomly assigned to Responses~A and~B for every example. We report win rates over 100 pairs together with tie rates, and Figures~\ref{fig:placeholder} and~\ref{fig:placeholder2} show that DWA-KD consistently outperforms all baselines in instruction-following quality.

\subsection{Representation Similarity between Teacher and Student Models}
\begin{figure}[h!]
  \centering

  \begin{subfigure}{\linewidth}
    \centering
    \includegraphics[width=\linewidth,trim=0 0 0 0,clip]{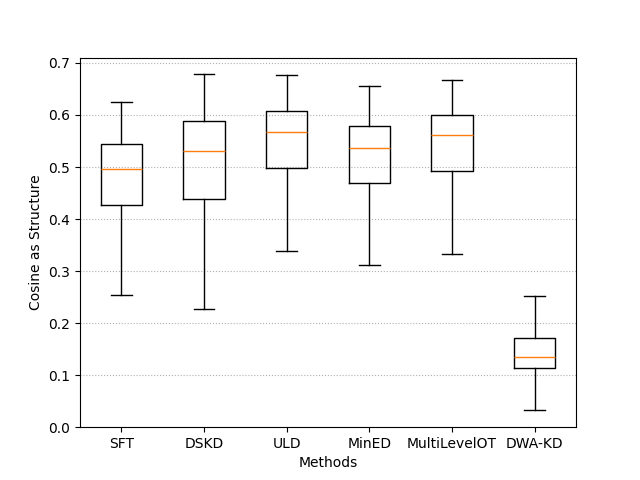}
  \end{subfigure}

   \vspace{-1em} 

  \begin{subfigure}{\linewidth}
    \centering
    \includegraphics[width=\linewidth,trim=0 0 0 0,clip]{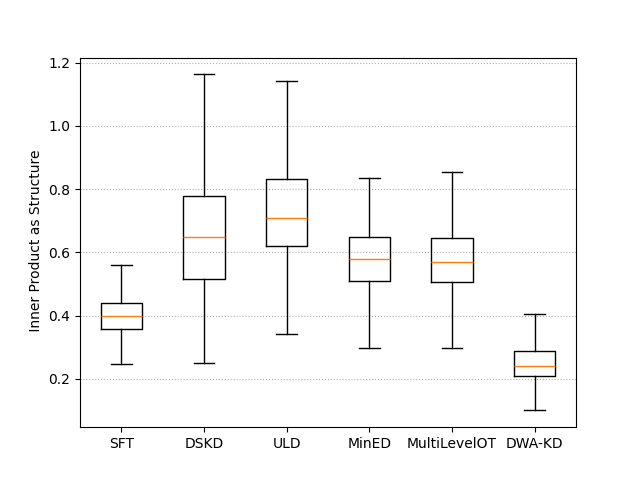}
  \end{subfigure}

  \caption{Teacher–student representation structure distance
  (cosine and normalized inner product; lower is better) for the Qwen-1.5--1.8B $\to$ GPT2-120M pair.}
  \label{fig:ts-structure-distance-tiny}
\end{figure}

To conduct an empirical study to compare token-level response patterns between teacher and student models despite differing hidden state dimensions, we adopt the approach from \cite{zhang2024dual}. For a given sentence with n tokens, we construct n × n structure matrices using cosine similarity and normalized inner products of the models’ hidden states. Lower values indicate that the student’s representations are closer to the teacher’s. We compute these distances over 100 training sentences and present results as box plots for the teacher–student pair Mistral-7B → TinyLlama-1.1B. Across Figure~\ref{fig:ts-structure-distance-tiny}, DWA-KD attains the smallest median distances, significantly smaller than others', on both structure matrices—cosine and normalized inner product—indicating that DWA-KD more faithfully preserves the teacher’s token-to-token representation structure than all other methods.

\subsection{Ablation study}
\paragraph{Impact of Each Component} To understand the contribution of each module we introduced in our DWA-KD framework, we conducted an ablation study comparing 4 configurations: (1) using only entropy weighting, (2) using only the DTW alignment loss, (3) using only the DTW alignment loss with banding and (4) removing the gating mechanism we introduced in (\ref{eq:gate_teacher_to_student}). Table \ref{tab:ctkd_ablations} shows that each component contributes positively to the overall performance. Incorporating the DTW loss consistently improves alignment quality, while entropy-based weighting enhances token-level selectivity. Adding the adaptive banding further refines sequence matching, leading to the best overall performance when all modules are combined in the full DWA-KD framework.
\begin{table}[t]
\centering
\footnotesize
\renewcommand{\arraystretch}{1} 
\setlength{\tabcolsep}{1pt}     

\begin{tabular}{l|ccccc}
\toprule
\textbf{Methods} & \textbf{Dolly} & \textbf{SelfInst} & \textbf{Vicuna} & \textbf{S-NI} & \textbf{Avg.} \\
\specialrule{1.0pt}{1.0pt}{1.0pt}
\midrule
\multicolumn{6}{c}{\textit{Qwen-1.5--1.8B $\rightarrow$ GPT2-120M}}\\
\midrule
$\mathcal{L}_{DSKD}$ & 24.26 & 10.07 & 15.25 & 17.15 & 16.68 \\
\hspace{1em}w/ EW & 23.99 & 10.83 & 15.65 & 18.44 & 17.23 \\
\hspace{1em}w/ DTW & 23.76 & \textbf{11.26} & 15.25 & 18.93 & 17.30 \\
\hspace{1em}w/ bDTW & 24.05 & 10.77 & 15.75 & 18.42 & 17.25 \\
\hspace{1em}w/o gate & 23.94 & 10.89 & \textbf{15.82} & 19.36 & 17.50 \\
\textbf{DWA-KD} & \textbf{24.29} & 11.15 & 15.65 & \textbf{19.63} & \textbf{17.68} \\
\midrule
\multicolumn{6}{c}{\textit{Mistral-7B $\rightarrow$ TinyLLaMA-1.1B}}\\
\midrule
$\mathcal{L}_{DSKD}$ & 26.28 & 17.19 & \textbf{18.74} & 31.93 & 23.54 \\
\hspace{1em}w/ EW & 26.36 & 18.01 & 17.97 & 34.77 & 24.28 \\
\hspace{1em}w/ DTW & 26.36 & 18.35 & 17.72 & 32.39 & 23.71 \\
\hspace{1em}w/ bDTW & \textbf{26.67} & 18.84 & 17.76 & 33.17 & 24.11 \\
\hspace{1em}w/o gate & 26.47 & 18.77 & 16.91 & \textbf{37.98} & 25.03 \\
\textbf{DWA-KD} & 26.38 & \textbf{19.62} & 17.63 & 36.59 & \textbf{25.06} \\
\bottomrule
\end{tabular}%

\caption{Ablation results on instruction-following benchmarks. ``Avg.'' denotes the mean over the listed datasets. EW: entropy-only token weights; DTW: Soft-DTW loss; bDTW: Soft-DTW with attention-entropy banding; gate: applying the teacher$\to$student max-probability gate.}

\label{tab:ctkd_ablations}
\end{table}




\section{Conclusion}
We propose \textbf{DWA-KD}, a knowledge distillation framework that aligns teacher and student at both the token and sequence levels. At the token level, DWA-KD extends DSKD with importance weights to emphasize informative tokens during distillation. At the sequence level, it applies DTW to both embedding and final hidden-state sequences, aligning lexical and semantic trajectories across tokenizers. Experiments show that DWA-KD improves teacher–student alignment and consistently boosts student performance across diverse evaluation settings. Future work will explore multi-teacher distillation for stronger knowledge integration.

\section{Limitation}
While DWA-KD effectively improves knowledge transfer across heterogeneous tokenizers, several limitations remain. First, the Soft-DTW loss introduces quadratic complexity in sequence length, which can increase computational and memory overhead. Second, the approach relies on a stable teacher–student projection, and poor initialization may weaken early-stage supervision. Moreover, our work was conducted under limited computational budgets, which constrained the scale and diversity of experimentation. We view these limitations as opportunities for further development and broader evaluation, particularly with stronger projection/calibration schemes under tight compute.

\section*{Acknowledgments}

This project is funded by Vietnam National Foundation for Science and Technology Development (NAFOSTED) under grant number 102.05-2025.16.

\bibliography{custom}

\newpage
\onecolumn
\appendix

\section{Experimental Details}
\label{sec:appendix_B}
\subsection{Data}
All datasets in our experiments are preprocessed following \citet{wan2024knowledge}. For training, we use the Databricks-Dolly-
15k dataset, which contains human-written instruction–response pairs and consists of 11435 samples for training, 1000 for
validation, and 500 for testing. To evaluate the generalization ability of our models beyond the Dolly domain, we employ 3 additional datasets:
\begin{itemize}
    \item SelfIns \cite{wang2023selfinstruct}: A user-oriented instruction-following set contains 252 samples, covering a diverse range of practical tasks such as email composition, social-media postings, entertainment prompts or programming assistance.
    \item VicunaEval \cite{chiang2023vicuna}: The 80 challenging questions used in the Vicuna evaluation, spanning 9 categories such as writing, roleplay, math, coding, and knowledge.
    \item S-NI \cite{wang2022benchmarking}: A large benchmark of NLP tasks and their natural language instructions. The test set of SuperNatural-Instructions consisting of 9K samples ranging from 119 tasks. Following \citet{peng2023instructiontuning}, we split the set into 3 subsets whose ground response lengths lie in [0, 5], [6, 10] and [11, $+\infty$], and use the [11, $+\infty$] subset comprising
1694 samples.
\end{itemize}

\subsection{Baselines}
We compare our method against state-of-the-art knowledge distillation techniques specifically designed to address discrepancies in tokenizers and vocabularies. These baselines provide a comprehensive framework for evaluating our method’s effectiveness to handle tokenizer and vocabulary disparities:
\begin{itemize}
    \item ULD \cite{boizard2024cross}: employs a Wasserstein distance-based loss to replace KL divergence, enabling distillation across different architectures and tokenizers.
    \item MinED \cite{wan2024knowledge}: introduces a method based on Dynamic Time Warping (DTW) for sequence alignment, which aligns the logits between two token lists of the same sentence tokenized by different tokenizers.
    \item DSKD \cite{zhang2024dual}: unifies the output spaces of the teacher and student distributions by projecting the output hidden states of teacher/student to the representation spaces of student/teacher, overcome limitations in similarity caused by their differing output spaces.
    \item MultiLevelOT \cite{cui2025multilevelot}: a cross-tokenizer knowledge distillation approach that leverages both sequence-aware token-level and sequence-level optimal transport for comprehensive distribution matching.
\end{itemize}
\subsection{Training and Evaluation}
\begin{table*}[!h]
\centering
\small
\renewcommand{\arraystretch}{1.2}
\setlength{\tabcolsep}{3pt}
\begin{tabular}{l|ccc|cc|ccc}
\hline
\textbf{Settings} & \multicolumn{3}{c|}{\textbf{KD with Qwen1.5-1.8B}} & \multicolumn{2}{c|}{\textbf{KD with Mistral7B}} & \multicolumn{3}{c}{\textbf{KD with Qwen2.5}} \\ 
\cline{2-9}
 & GPT2-120M & GPT2-340M & Qwen1.5 & TinyLLaMA & Mistral7B & GPT2-1.5B & OPT2.7B & Qwen2.5 \\ 
\hline
Epoch & 20 & 20 & 15 & 15 & 15 & 15 & 15 & 15 \\ 
LR & $5\times10^{-4}$ & $5\times10^{-4}$ & $2\times10^{-5}$ & $1\times10^{-3}$ & $1\times10^{-3}$ & $1\times10^{-3}$ & $1\times10^{-3}$ & $1\times10^{-3}$ \\ 
Projector LR & $1\times10^{-3}$ & $1\times10^{-3}$ & $1\times10^{-3}$ & $1\times10^{-3}$ & $1\times10^{-3}$ & $1\times10^{-3}$ & $1\times10^{-3}$ & $1\times10^{-3}$ \\ 
Batch Size & 8 & 8 & 8 & 8 & 8 & 8 & 8 & 8 \\ 
LR Scheduler & Cosine & Cosine & Cosine & Cosine & Cosine & Cosine & Cosine & Cosine \\ 
Fine-Tuning Method & Full & Full & Full & LoRA & LoRA & LoRA & LoRA & LoRA \\ 
LoRA Rank & N/A & N/A & N/A & 256 & 256 & 256 & 256 & 256 \\ 
LoRA Alpha & N/A & N/A & N/A & 8 & 8 & 8 & 8 & 8 \\ 
LoRA Dropout & N/A & N/A & N/A & 0.1 & 0.1 & 0.1 & 0.1 & 0.1 \\ 
\hline
\end{tabular}
\caption{Detailed training configurations of KD with Qwen1.5-1.8B, Mistral7B, and Qwen2.5.}
\label{tab:kd_configs}
\end{table*}
We list the detailed training configurations for all models we trained in Table \ref{tab:kd_configs}. For evaluation, we used random sampling to decode the responses from all models. We set the decoding temperature and top\_p to 1.0. We then generate the responses with 5 random seeds and report the averaged ROUGE-L score \cite{lin2004rouge} of each seed. Besides, we also evaluate models using GPT-4.1 judgments. We randomly sample 100 instructions from the Dolly test set and generate one response per model using the same decoding settings (temperature and top-p = 1.0). The GPT-4.1 API is then prompted with a fixed evaluation template (Figure \ref{fig:gpt4}) to perform pairwise comparisons. To reduce position bias \cite{zheng2023judging}, the two model outputs are randomly ordered as Response A and Response B for each case. We report the proportion of wins and ties across the 100 comparisons. In all experiments, we use fixed DTW banding hyperparameters without tuning; see Table~\ref{tab:banding-hparams} for values.
\begin{table}[!h]
\centering
\setlength{\tabcolsep}{6pt}
\begin{tabular}{lp{7.8cm}l}
\toprule
\textbf{Symbol} & \textbf{Description} & \textbf{Value} \\
\midrule
$b$ & Base band width (in tokens) & 5 \\
$\beta$ & Entropy sensitivity & 2.0 \\
$\alpha$ & Blend between soft center and linear diagonal & 0.7 \\
$\lambda$ & Additive cost outside the band & 1.0 \\
\bottomrule
\end{tabular}
\caption{Hyperparameters used for soft-DTW banding.}
\label{tab:banding-hparams}
\end{table}

\section{Representation Similarity between Teacher and Student Model}
\label{sec:appendix_C}
Following \citet{zhang2024dual}, we empirically evaluate how well the student’s token-level response patterns align with those of the teacher in a realistic KD setting. As the two models differ in hidden dimensionality, direct token-wise comparison of hidden states is infeasible. Instead, given the same input sentence, we compare the structure of token-to-token responses by constructing $n \times n$ structure matrices—computed via cosine similarity and normalized inner products between the models’ output hidden states:
\[
\mathcal{M}_{\text{cosine}}(i, j) = 
\frac{h_i^{\top} h_j}{\|h_i\| \|h_j\|} \in \mathbb{R}^{n \times n}
\]

\[
\mathcal{M}_{\text{prod}}(i, j) = 
\frac{h_i^{\top} h_j}{\sum_k h_i^{\top} h_k} \in \mathbb{R}^{n \times n}
\]
where $\mathcal{M}_{\text{prod}}$ and $\mathcal{M}_{\text{prod}}$ are structure matrices calculated by cosine and normalized inner-product between output hidden states, respectively. Then we calculate the L1 distance between the matrices of the
student and the teacher:
\[
\mathcal{D}_{\text{cosine}} = 
\sum_{i=1}^{n} \sum_{j=1}^{n} 
\left| \mathcal{M}_{\text{cosine}}^{t}(i, j) - \mathcal{M}_{\text{cosine}}^{s}(i, j) \right|
\]

\[
\mathcal{D}_{\text{prod}} = 
\sum_{i=1}^{n} \sum_{j=1}^{n} 
\left| \mathcal{M}_{\text{prod}}^{t}(i, j) - \mathcal{M}_{\text{prod}}^{s}(i, j) \right|
\]
Lower values indicate closer alignment between student and teacher representations. We compute these distances on 1,000 training sentences and visualize them using box plots. Two teacher–student pairs are evaluated: Mistral-7B → TinyLLaMA-1.1B. As shown in figures \ref{fig:ts-structure-distance-tiny}, DWA-KD achieves the lowest median L1 distances on both cosine and product-based structure matrices, showing that it better preserves the teacher’s token-level structure.

\section{Attention matrix in DTW banding}
\label{sec:appendix_D}
Following \citet{zhang2024dual}, we use a \textit{cross-model attention (CMA)} matrix that automatically learns the alignment between tokens of the student sequence consist of S tokens and teacher sequence consist of T tokens. Specifically, the student's input embeddings $e^s_{1:S}$ and target embeddings $e^s_{2:S+1}$ are concatenated along the last dimension, and project them through a query projector $P_q$:
\[
Q = P_q([e^s_{1:S}; e^s_{2:S+1}]; \theta_P^q) \in \mathbb{R}^{S \times 2D},
\]
For the teacher model, the key and value vectors are obtained from its embeddings and output hidden states:
\[
K = N([e^t_{1:T}; e^t_{2:T+1}]) \in \mathbb{R}^{T \times 2D}, \]
\[
V = P_v(N(e^t_{2:T+1}) + N(h^t_{1:T}); \theta_P^v) \in \mathbb{R}^{T \times d},
\]
where $N(x) = x / \mathrm{std}(x)$ denotes normalization by the standard deviation, which promotes faster convergence.
The attention matrix is then computed as:
\[
A = \mathrm{softmax}\!\left(\frac{QK^\top}{\sqrt{2D}}\right) \in \mathbb{R}^{T \times S}.
\]
The attention-based center in Eq.~\ref{eq:center} leverages the cross-model attention matrix $A$ to capture semantic alignment between student and teacher tokens. The term $\sum_{j} j A_{i,j}$ reflects a content-informed correspondence, indicating where the teacher's information most strongly aligns with the student token $i$. In contrast, the linear center $i \cdot \tfrac{T}{S}$ encodes an absolute positional prior assuming uniform progression. Blending the two allows the band center to incorporate both semantic and positional alignment cues, yielding a more adaptive and stable alignment.

\section{Additional Experiments}
\label{appdx: additional}
\subsection{Comparing with stronger CTKD baselines}
To better position DWA-KD against more recent cross-tokenizer distillation approaches suggested by reviewers, we implemented and evaluated two additional baselines: ALM \citep{minixhofer2025universalcrosstokenizerdistillationapproximate} and DSKDv2 \citep{zhang2025dualspaceframeworkgeneralknowledge}, following their official descriptions/implementations under the same training recipe used throughout the paper (same data, epochs, optimizer settings, and evaluation protocol). Across representative teacher–student pair (\textbf{Qwen-1.5-1.8B} $\rightarrow$ \textbf{GPT2-120M}), DWA-KD remains consistently competitive and achieves higher average performance.

\begin{table}[!ht]
\centering
\small
\begin{tabular}{lcccc|c}
\toprule
Method & Dolly & SelfInst & Vicuna & S-NI & Avg. \\
\midrule
\multicolumn{6}{c}{\textbf{Qwen-1.5-1.8B} $\rightarrow$ \textbf{GPT2-120M}}\\
\midrule
Teacher   & 28.23 & 19.58 & 19.59 & 34.36 & 25.44 \\
SFT       & 23.78 &  6.78 & 17.04 &  7.81 & 13.85 \\
ALM       & 20.86 & 10.65 & 14.76 & 17.76 & 16.01 \\
DSKD      & 24.26 & 10.07 & 15.25 & 17.15 & 16.68 \\
DWA-KD    & 24.29 & 11.15 & 15.64 & 19.63 & 17.68 \\
DSKDv2    & 24.37 & 10.89 & 15.52 & 20.37 & 17.78 \\
DWA-KDv2  & 24.66 & 10.96 & 15.21 & 20.71 & 17.89 \\
\bottomrule
\end{tabular}
\caption{Comparison with DSKDv2 and ALM on a representative teacher-student pair.}
\label{tab:dskdv2_extra}
\end{table}
\subsection{Efficiency and resource overhead}
Since DWA-KD includes a sequence-level Soft-DTW objective, we additionally report step time and GPU memory usage measured under the same sequence length regime as our main experiments. Table \ref{tab:efficiency_extra} shows that DWA-KD incurs only a modest overhead compared to common GPU-based CTKD baselines (DSKD, MinED, ULD) at batch size 4, while remaining substantially more efficient than CPU-based CDM in its supported batch-size-1 setting.

\begin{table}[!ht]
\centering
\small
\begin{tabular}{lccc}
\toprule
Method & Avg Mem (GB) & Peak Mem (GB) & Step time (s) \\
\midrule
\multicolumn{4}{c}{\textbf{Batch size 4}}\\
\midrule
DSKD   & 20.11 & 26.38 & 0.35 \\
MinED  & 19.63 & 22.29 & 0.42 \\
ULD    & 19.63 & 27.17 & 0.44 \\
DWA-KD & 20.17 & 29.92 & 0.45 \\
\midrule
\multicolumn{4}{c}{\textbf{Batch size 1}}\\
\midrule
CDM*   & 22.61 & 24.64 & 1.01 \\
DWA-KD & 20.16 & 22.36 & 0.30 \\
\bottomrule
\end{tabular}
\caption{Efficiency measurements. *CDM uses the authors' original implementation which does not support batch sizes $>1$, hence the batch-size-1 comparison reflects its practical regime.}
\label{tab:efficiency_extra}
\end{table}
\section{Prompt for GPT-4.1 evaluation}
\begin{figure}[!h]
\centering
\begin{tcolorbox}[colback=promptbg, colframe=black, boxrule=0.5pt,
                  left=3pt,right=3pt,top=3pt,bottom=3pt,
                  width=1\columnwidth] 
\scriptsize 
\raggedright
\textbf{Please act as an impartial judge} and compare the quality of response A and response B
provided by two AI assistants to the user question displayed below. Your evaluation should
consider factors such as the helpfulness, relevance, accuracy, depth, creativity, and level of
detail of the response.

\medskip
Just tell me which response you think is better:\\
  - If A is significantly better than B, just answer me ``A'';\\
  - If B is significantly better than A, just answer me ``B'';\\
  - If A and B have similar quality (both good or both wrong), just answer me ``Tied''.\\

\medskip
\textbf{[Question]}\\
\{question or instruction\}

\medskip
\textbf{[Response A]}\\
\{response A\}

\medskip
\textbf{[Response B]}\\
\{response B\}
\end{tcolorbox}
\caption{Prompt for GPT-4.1 evaluation.}
\label{fig:gpt4}
\vspace{-2mm}
\end{figure}

\end{document}